\documentclass[10pt,twocolumn,letterpaper]{article}

\usepackage{wacv}              

\usepackage{graphicx}
\usepackage{amsmath}
\usepackage{amssymb}
\usepackage{booktabs}
\usepackage{graphicx}
\usepackage{amsmath}
\usepackage{amssymb}
\usepackage{booktabs}
\usepackage{tikz}
\usepackage{bbding}
\usepackage{graphicx}
\usepackage{amsmath}
\usepackage{amssymb}
\usepackage{booktabs}
\usepackage{multirow} 
\usepackage{array}
\usepackage{bbm}
\newcommand{\greencheck}{{\color{green}\Checkmark}}
\newcommand{\redcross}{{\color{red}\XSolidBrush}}

\usepackage[pagebackref,breaklinks,colorlinks]{hyperref}

\usepackage[capitalize]{cleveref}
\crefname{section}{Sec.}{Secs.}
\Crefname{section}{Section}{Sections}
\Crefname{table}{Table}{Tables}
\crefname{table}{Tab.}{Tabs.}

\def\wacvPaperID{*****} 
\def\confName{WACV}
\def\confYear{2025}

\begin{document}


\title{Metric for Evaluating Performance of Reference-Free Demorphing Methods}

\author{Nitish Shukla\\
Michigan State University\\
{\tt\small shuklan3@msu.edu}
\and
Arun Ross\\
Michigan State University\\
{\tt\small rossarun@msu.edu}
}
\maketitle
\begin{abstract}
   A facial morph is an image created by combining two (or more) face images pertaining to two (or more) distinct identities. Reference-free face demorphing inverts the process and tries to recover the face images constituting a facial morph without using any other information. However, there is no consensus on the evaluation metrics to be used to evaluate and compare such demorphing techniques. In this paper, we first analyze the shortcomings of the demorphing metrics currently used in the literature. We then propose a new metric called biometrically cross-weighted IQA that overcomes these issues and extensively benchmark current methods on the proposed metric to show its efficacy. Experiments on three existing demorphing methods and six datasets on two commonly used face matchers validate the efficacy of our proposed metric.

\end{abstract}

\section{Introduction}
\begin{figure}
    \centering
    \includegraphics[width=\linewidth]{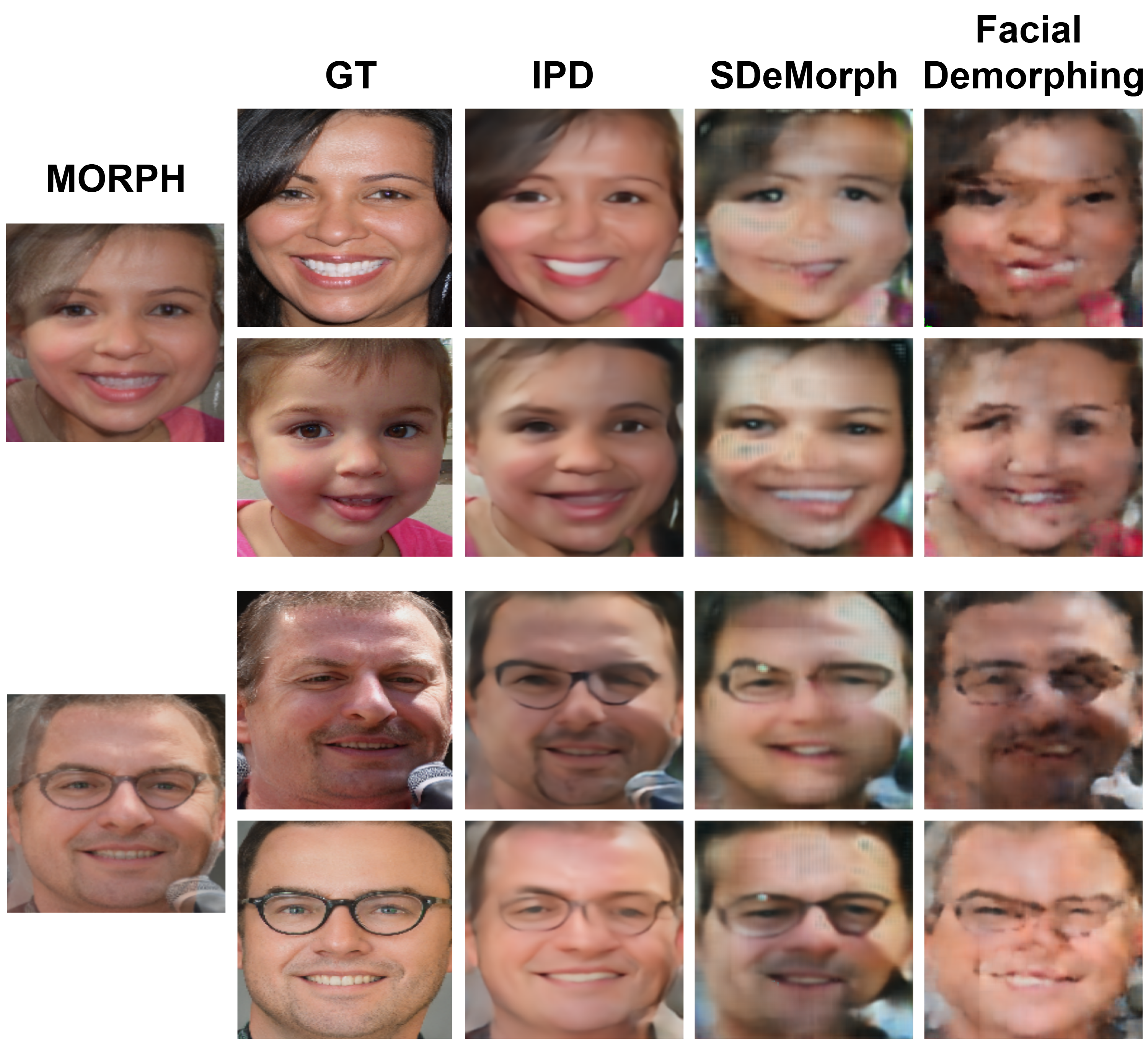}
    \caption{Single-image reference-free demorphing: The MORPH image is created by blending ground-truth (GT) face images. Reconstructions are produced using Identity Preserving Decomposition (IPD) \cite{ref66}, SDeMorph \cite{ref18}, and Facial Demorphing \cite{ref51}. Among these, IPD achieves the best visual quality, which aligns with the score generated by our proposed metric.}
    \label{fig:enter-label}
\end{figure}
\label{sec:intro}
A face morph is created by blending two or more face images pertaining to two (or more) distinct individuals. The goal is to create a face image that can match all component identities with respect to an automated face matcher or a human face examiner \cite{ref116,ref27}.  As a result, facial morphs can be maliciously used to  covertly allow multiple individuals to share a single ID document such as a passport \cite{ref11,ref10}. Typically, a successful morph utilizes identities having similar characteristics, \textit{e.g.}, similar race, age group, ethnicity or visual facial features. Historically, morphs were created by extracting landmarks from component images, aligning the images using these landmarks, and then blending them \cite{ref117,ref118}. However, recently, end-to-end deep-learning based techniques are being used to generate morphs, wherein facial alignment and blending are implicitly accomplished by the neural network. In particular, generative models like GANs\cite{ref32} and diffusion models\cite{ref108} have been successfully used in this endeavor \cite{ref107,ref119,ref9,ref120,ref121}.
\begin{table*}[]
    \centering
        \caption{Comparison of metrics used by existing demorphing  methods under various assumptions.}
    \label{tab:my_label}
    \resizebox{\linewidth}{!}{
    \begin{tabular}{|c|c|c|c|c|c|c|}
         \hline
         Method&Venue&Scenario& \begin{tabular}{c}
              Assume train/test \\ morph technique  
         \end{tabular} & TMR @10\% FMR & RA &IQA\\
         \hline
          Facial Demorphing \cite{ref51} & IJCB, 2022 & 3 &  \greencheck & \greencheck &\redcross &\redcross \\
          SDeMorph \cite{ref18} & IJCB, 2023 & 1 &  \greencheck & \redcross &\greencheck &\greencheck \\
          Identity Preserving Decomposition(IPD)\cite{ref66} & IJCB, 2024 & 1 &  \greencheck & \redcross &\greencheck &\greencheck \\
          
          \hline
    \end{tabular}}

\end{table*}

Morph Attack Detection (MAD) is, therefore, crucial for the integrity and security of face-based biometric systems. MAD can be broadly categorized into reference-based (RB) differential image methods\cite{ref36,ref53,ref16} and single image reference-free (RF) methods\cite{ref90,ref122,ref123}. While MAD techniques can detect morphs, they do not reveal any information about the constituent images used to create the morph. Demorphing addresses this issue. In this paper, we focus on reference-free demorphing.

Reference-free face demorphing is an ill-posed inverse problem due to lack of constraints in image space as well as absence of prior information such as the morph technique used.  Indeed, given a morphed image, an infinite number of decompositions is possible, making it challenging to reliably recover the constituent face images. While existing reference-free methods perform demorphing with varying degrees of success, there is no consensus on the evaluation metrics used. In \cite{ref51}, the authors used True Match Rate (TMR) at 10\% False Match Rate (FMR) to evaluate the performance of their demorphing method. However, in \cite{ref18,ref66}, authors used Restoration Accuracy (RA) and commonly used Image Quality Assessment (IQA) metrics like Structural Similarity Index Measure (SSIM) and Peak Signal-to-Noise Ratio (PSNR) as performance metrics.

True Match Rate (TMR) is defined as the percentage of times a biometric system correctly identifies a genuine match between a user's biometric sample and their enrolled data. The threshold of the match is computed through False Match Rate (FMR). Thus, TMR @10\% FMR represents the percentage of instances where the output is correctly matched to the ground truth, allowing for a 10\% rate of incorrect matches.
Restoration Accuracy (RA) is similar to TMR with  the difference that the threshold is set to a fixed value. On the other hand, IQA metrics work in RGB pixel domain. SSIM is used to measure similarity between two images using the structural information (luminance, contrast) of the pixels in regions of the image. Finally, PSNR is used to compare image reconstruction quality. It is defined as the ratio between the maximum possible power of a signal and the power of distorting noise that affects the quality of its reconstruction.

However, all of the metrics described have several drawbacks when used in the context of demorphing. While TMR and RA only focus on the biometric information in reconstructed images and ignore the quality of output images, PSNR and SSIM ignore the biometric aspect. Therefore, there is a need for a comprehensive metric that addresses these limitations. In this paper, we propose a novel evaluation metric for reference-free demorphing methods that balances biometric utility with image quality while penalizing trivial demorphing outputs.  In summary, our contributions are as follows:
\begin{itemize}
    \item We benchmark existing reference-free demorphing techniques\footnote{To the best of our knowledge, these are the only open-source reference-free demorphing methods in the literature.} \cite{ref18,ref66,ref51} under a standardized training protocol and evaluate them based on three existing metrics.
    \item We propose a new evaluation metric called \textit{biometrically cross weighted IQA}. Our proposed metric overcomes the flaws in existing metrics and is observed to be applicable across different datasets and face matchers.

\end{itemize}

The rest of the paper is organized as follows: In Section \ref{demorphing}, we discuss facial demorphing and setup the problem. Section 
 \ref{sec:previous} discusses existing work and their limitations.  In Section \ref{ref:metrics}, we discuss existing metrics and introduce our proposed evaluation metric. In Section \ref{sec:datasets} we discuss the datasets used in this work. In Section \ref{sec:experiments} we present the experiments conducted and discuss the results in Section \ref{sec:results}. We summarize the work in Section \ref{sec:conclusion}.

\section{Face Demorphing}
\label{demorphing}

For the sake of consistency across all methods evaluated, we denote the morph image as $x$, the constituent face images as $i_1, i_2$ such that: 
\begin{equation}
    x=\mathcal{M}(i_1,i_2)
\end{equation}
where, $\mathcal{M}$ is the morphing operator. The goal of $\mathcal{M}$ is to ensure that
$\mathcal{B}(x,i_k)>\tau$, $k\in \{1,2\}$, with respect to a face recognition software $\mathcal{B}$ and similarity threshold $\tau$.  A demorphing operator, denoted as $\mathcal{DM}$ acts upon $x$ and aims to recover the constituent face images,
\begin{equation}
    o_1,o_2=\mathcal{DM}(x)
\end{equation}
satisfying the following conditions: 

\begin{equation}
\label{eq3}
\mathcal{B}(o_1,o_2)<\theta
\end{equation}

\begin{equation}
\label{eq4}
    \min_{j\in\{1,2\}} \max_{\substack{k \in \{1,2\} \\ k \neq j}} \{\ \ \mathcal{B}(o_j,i_k),\mathcal{B}(o_j,i_j)\ \ \} >\epsilon
\end{equation}
Eqn. (\ref{eq3}) enforces the reconstructed outputs to look dissimilar among themselves to avoid morph replication, while Eqn. (\ref{eq4}) enforces each output to align with its corresponding ground truth image.

In the literature, demorphing has been performed under various scenarios primarily based on the protocol used to define the train and test morphs\cite{ref66,dcgan}.  Let $\mathcal{X}_{train}$, $\mathcal{Y}_{train}$ denote the training data corresponding to the morphs and constituent face images, respectively. Assume that all morph images in $\mathcal{X}_{train}$ are necessarily generated only using the identities in $\mathcal{Y}_{train}$.  
\begin{equation}
    \forall x\in \mathcal{X}_{train}, \mathcal{M}(y_1,y_2)=x \Rightarrow \{y_1,y_2\}\in \mathcal{Y}_{train} 
\end{equation}
where, $\mathcal{M}$ is the morphing operator. Test sets $\mathcal{X}_{test}$, $\mathcal{Y}_{test}$ are defined similarly. 
The following three scenarios are envisioned:
1) Train and test morphs are generated from the same pool of identities. Although the images used to create the train and test morphs are identical, the same pair of images is not selected for both i.e., given a test morph $x$ created from face images $i_1$ and $i_2$, scenario 1 asserts that $i_1 \in \mathcal{Y}_{train}$ and $i_2 \in \mathcal{Y}_{train}$.
2) Test morphs are created using one face image from an identity seen during training and one from an unseen identity i.e., scenario 2 asserts that either $i_1 \in \mathcal{Y}_{train}$ or  $i_2 \in \mathcal{Y}_{train}$ but not both. 3) Train and test morphs are generated from a disjoint pool of face images (and disjoint identities) i.e., the identities involved in creating training morphs do not participate in the creation of testing morphs and vice-versa. In other words, $\mathcal{Y}_{train} \cap \mathcal{Y}_{test} =\phi$. Note that the datasets used in this paper have one face image per identity (only neutral face images from FRLL dataset are used to create the morphs). So in some places, the terms “images” and “identities” have been used interchangeably.

These conditions are formally defined as:
\begin{equation}
\begin{aligned}
    \text{Scenario 1)} \ \ \ \ \  \ \mathcal{Y}_{test}\subseteq\mathcal{Y}_{train} ,& \ \mathcal{X}_{train} \cap \mathcal{X}_{test}=\phi \\
\text{Scenario 2)}\  \mathcal{Y}_{test}\cap\mathcal{Y}_{train}\neq \phi ,& \ \mathcal{X}_{train} \cap \mathcal{X}_{test}=\phi \\
    \text{Scenario 3})\ \mathcal{Y}_{test}\cap\mathcal{Y}_{train}\ =\ \  &\mathcal{X}_{train} \cap \mathcal{X}_{test}=\phi 
\end{aligned}
\label{eq:scenario}
\end{equation}

\section{Previous Work}
\label{sec:previous}

In this paper, we mainly focus on single-image reference-free demorphing. Reference-free (RF) demorphing only requires the morph image to reconstruct the constituent images, whereas reference-based (RB) demorphing requires both the morph image and one of the constituent images to recover the other.  RF demorphing is a relatively recent topic due to its inherent complexities. In an initial work \cite{ref51},  authors used GANs to recover the constituent images from the morph image. Their method employs an image-to-image generator and three markovian discriminators. Their model is trained using a combination of patch-based loss  and cross-road loss \cite{ref104} along with the standard GAN adversarial loss. While the authors assumed  scenario 3, their method had two main limitations: i) their method tends to replicate the input morph as its outputs, a problem referred to as \textit{morph replication}\cite{dcgan}, and ii) they assumed that train
and test morphs were generated using the same technique. They also used TMR @ 10\% FMR to evaluate their method which is not a suitable metric for evaluating  demorphing methods (see section \ref{sec:experiments}). In \cite{ref18}, the authors used diffusion to first iteratively add noise to the morph image until the image degenerates to a noisy artifact. In the backward process, their method tries to recover the constituent face images by learning the noise added at each timestep in the forward process. While their method performed very well in terms of Restoration Accuracy, the method assumed scenario 1, i.e., the train and test morphs are created from the same pool of face images while making sure that the pair selected for creating the training morphs does not appear when creating the test morphs. This assumption is not realistic and hampers the usage of their method. In \cite{ref66}, authors introduce another demorphing technique; their method starts by decomposing the input morph into multiple unintelligible components using a decomposer network. A merger network weighs and combines these components to recover the constituent image used to create the morph. Their method also assumes scenario 1 thereby limiting its  practical applicability. In \cite{dcgan}, authors introduce a method for scenario 3 that tries to overcome issues in \cite{ref51} by imposing additional conditions on the generator. Their method uses a perceptual image encoder to encode the morph; this encoded information is injected to the intermediate layers of the generator along with the morph image to effectively guide the generator and overcome the problem of morph-replication.

\begin{figure}
    \centering
    \includegraphics[width=\linewidth]{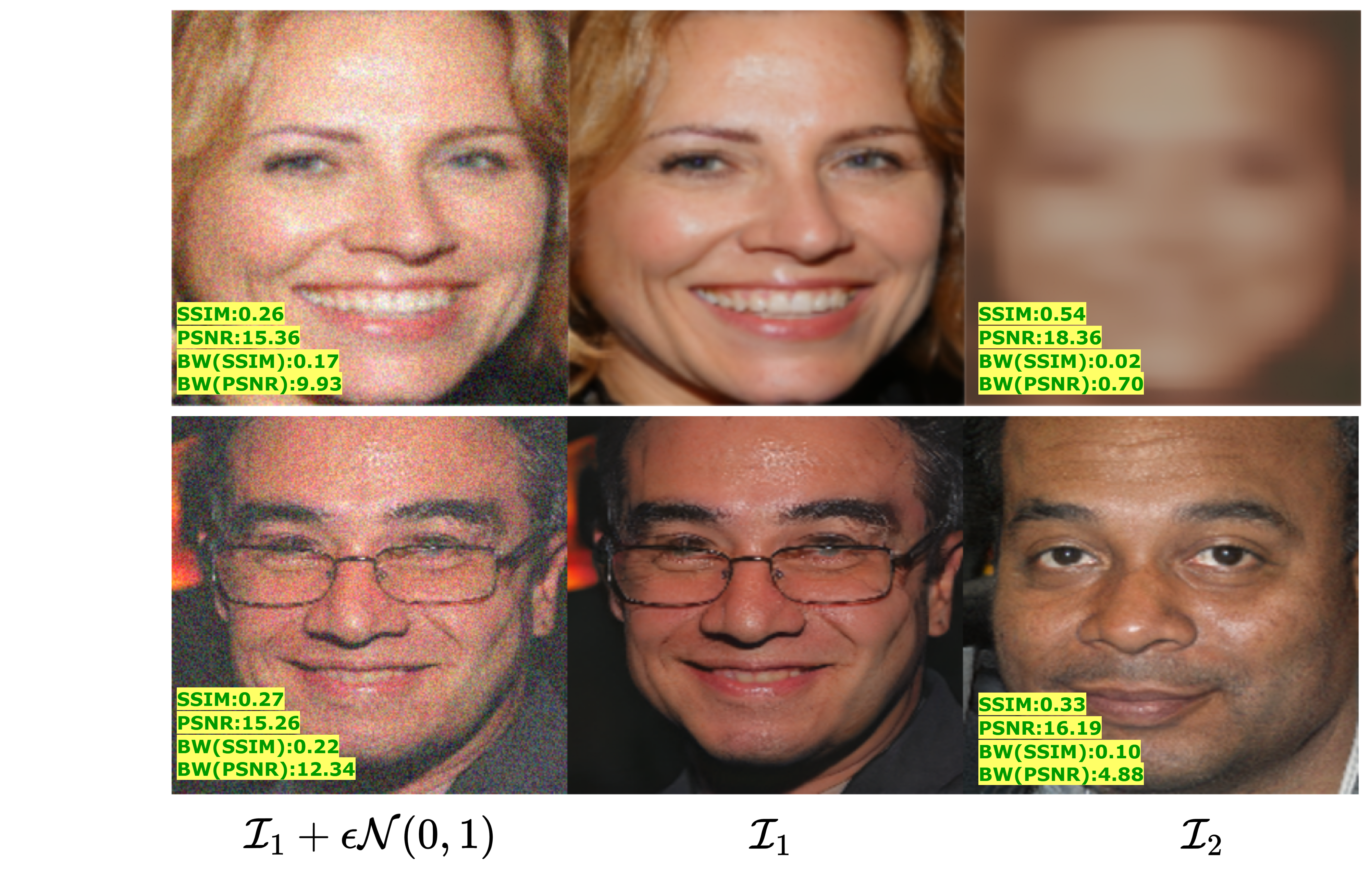}
    \caption{Comparison of the efficacy of SSIM and PSNR with that of the proposed metric for demorphing ($\epsilon=0.3$). (Top) The middle image is more structurally similar to the blurred image on right compared to the noisy image of the same subject on the left. (Bottom) $\mathcal{I}_1$  is more similar to $\mathcal{I}_2$, which belong to different subjects, compared to the noisy version of $\mathcal{I}_1$. Our proposed metric correctly balances the identity and image quality to produce consistent scores. }
    \label{fig:ssim-psnr}
\end{figure}

\begin{table*}[ht]
\centering
\caption{Evaluation of existing state-of-the-art demorphing methods (SDeMorph \cite{ref18}, Identity-Preserving Demorphing (IPD) \cite{ref66} and Facial Demorphing \cite{ref18}) under a common protocol. The scores reported are supplied by the original authors. }
\label{tab:results}
\resizebox{\linewidth}{!}{
\begin{tabular}{|c|c|c|c|c|c|c|c|c|c|c|c|c|c|c|c|c|c|c|c|c|c|}
\hline
\multirow{2}{*}{Dataset} & \multicolumn{7}{c|}{SDeMorph\cite{ref18}} & \multicolumn{7}{c|}{IPD\cite{ref66}} & \multicolumn{7}{c|}{Facial Demorphing\cite{ref51}} \\
\cline{2-22}
& \multirow{2}{*}{PSNR/SSIM}&\multicolumn{2}{c|}{Rest. Acc} & \multicolumn{2}{c|}{BW(SSIM)} & \multicolumn{2}{c|}{BW(PSNR)} & \multirow{2}{*}{PSNR/SSIM}&\multicolumn{2}{c|}{Rest. Acc} & \multicolumn{2}{c|}{BW(SSIM)} & \multicolumn{2}{c|}{BW(PSNR)} &\multirow{2}{*}{PSNR/SSIM}& \multicolumn{2}{c|}{Rest. Acc} & \multicolumn{2}{c|}{BW(SSIM)} & \multicolumn{2}{c|}{BW(PSNR)} \\
\cline{3-8}
\cline{10-15}
\cline{17-22}
& & AdaFace & ArcFace & AdaFace & ArcFace &  AdaFace & ArcFace & & AdaFace & ArcFace &  AdaFace & ArcFace & AdaFace & ArcFace&  & AdaFace & ArcFace& AdaFace & ArcFace& AdaFace & ArcFace \\
\hline
AMSL        & 8.99/0.34 & 0\% & 12.56\% & 0.11 & 0.16 & 2.76 & 4.24 & 9.32/0.38 & 0.18\% & 25.69\% & 0.17 & 0.26 & 4.14 & 6.28 & 9.68/0.46       &0.17\%       &0.45\%      & 0.11       &0.21       &2.39       &4.46     \\ \hline
OpenCV      & 9.53/0.37 & 0\% & 15.62\% & 0.12 & 0.19 & 3.2 & 4.88 & 10.37/0.42 & 1.89\% & 40.54\% & 0.21 & 0.32 & 5.3 & 7.98 &10.35/0.48       &0.23\%       &0.53\%      & 0.14       &0.24       &2.95       &5.21   \\ \hline
FaceMorpher & 9.60/0.37 & 0\% & 13.18\% & 0.12 & 0.19 & 3.22 & 4.97 & 10.27/0.41 & 1.43\% & 37.82\% & 0.21 & 0.32 & 5.29 & 9.95 & 10.44/0.47       &0.17\%       &0.51\%      & 0.13       &0.25       &2.89       &5.50   \\ \hline
WebMorph    & 9.45/0.37 & 0\% & 12.80\% & 0.12 & 0.18 & 3.16 & 4.68 & 9.63/0.39 & 0.31\% & 25.61\% & 0.16 & 0.25 & 4.1 & 6.16 & 10.2/0.48       &0.20\%       &0.50\%      & 0.13       &0.23       &2.82       &4.93 \\ \hline
MorDiff     & 8.96/0.34 & 0\% & 11.67\% & 0.11 & 0.17 & 2.98 & 4.29 & 9.91/0.40 & 3.88\% & 38.12\% & 0.22 & 0.33 & 5.53 & 8.12 &  10.13/0.47       &0.29\%       &0.62\%      & 0.16       &0.29       &3.33       &6.13   \\ \hline
StyleGAN    & 8.74/0.34 & 0\% & 0\% & 0.05 & 0.15 & 1.29 & 3.85 & 9.08/0.37 & 0.0\% & 16.22\% & 0.08 & 0.22 & 1.93 & 5.31 &  9.51/0.45       &0.00\%       &0.43\%      & 0.06       &0.19       &1.31       &4.13   \\ \hline
\textbf{Average} & \textbf{9.21/0.36} & \textbf{0\%} & \textbf{10.64\%} & \textbf{0.11} & \textbf{0.17} & \textbf{2.77} & \textbf{4.48} & \textbf{9.76/0.40} & \textbf{1.28\%} & \textbf{30.00\%} & \textbf{0.18} & \textbf{0.28} & \textbf{4.38} & \textbf{7.63} & \textbf{10.05/0.46} & \textbf{0.18\%} & \textbf{0.51\%} & \textbf{0.12} & \textbf{0.23} & \textbf{2.62} & \textbf{5.06} \\ \hline
\end{tabular}
}
\end{table*}

\begin{table*}[]
    \centering
        \caption{Average True Match Rate(TMR) @ 10\% False Match Rate (FMR) computed for the current state-of-the-art demorphing methods and compared with the proposed metric. The trivial solution, which simply replicates the morph as both of its outputs, achieves perfect TMR, rendering the metric uninformative. }
    \label{tab:tmr}
    \resizebox{0.5\textwidth}{!}{
    \begin{tabular}{|c|c|c|c|c|}
        \hline
          avg TMR @ 10\% FMR&SDeMorph\cite{ref18} & IPD\cite{ref66} & Facial Demorphing\cite{ref51} & trivial \\
          \cline{1-5}
         AMSL& 45.32\%& 72.36\%&22.13\% & 100\%\\
         OpenCV& 20.54\%& 68.74\%&21.72\% & 100\%\\

         FaceMorpher&57.01\% &67.69\% &20.05\% & 100\%\\
         WebMorph& 55.12\%&65.52\% &18.18\% & 100\%\\
         MorDiff& 53.76\%& 63.22\%&18.33\% & 100\%\\
         StyleGAN&30.54\% & 7.32\%& 7.02\%& 100\%\\
         
         \hline
         
    \end{tabular}}

\end{table*}
\section{Evaluation Metrics}
\label{ref:metrics}

In the literature, there are three primary metrics for evaluating face demorphing methods: True Match Rate (TMR), Restoration Accuracy (RA), and Image Quality metrics such as SSIM and PSNR. While TMR and RA focus solely on the biometric identity within the image, SSIM and PSNR disregard it entirely. Furthermore, because a morph image is inherently biometrically similar to the constituent images from which it was created, metrics that rely only on biometric features can  produce misleading results. 
\subsection{Evaluation Criterion}
Given a morph $x$, created using constituent face images $i_1,i_2$, the demorpher outputs $o_1$ and $o_2$. Since the outputs from the demorpher are unordered, we determine the correct pairing with the ground truth face images, $i_1$ and $i_2$,  by calculating the similarities for the two possible output-ground truth pairs. We use a face comparator (matcher) $\mathcal{B}$ to assess facial similarity. If the sum $\mathcal{B}(\text{$o_1$}, \text{$i_1$}) + \mathcal{B}(\text{$o_2$}, \text{$i_2$})$ is greater than $\mathcal{B}(\text{$o_1$}, \text{$i_2$}) + \mathcal{B}(\text{$o_2$}, \text{$i_1$})$, we deem ($o_1$, $i_1$) and ($o_2$, $i_2$) as the correct pairs; otherwise, we deem ($o_1$, $i_2$) and ($o_2$, $i_1$) as the correct pairs. For each $\text{$o$}_n$, we consider its associated pair $\text{$i$}_n$ as genuine, where $n \in$ \{1,2\}. The impostor score is computed by identifying the closest matching face in the original face image database, excluding the ground truth images used for creating that morph. Once the correct pairing between the outputs and ground truth images is done, computing PSNR and SSIM is straightforward.

\begin{equation}
IQA=0.5\times\max \left( 
\begin{aligned}
&(iqa(o_1,i_1)+iqa(o_2,i_2)), \\
&(iqa(o_1,i_2)+iqa(o_2,i_1)
\end{aligned}
\right)
\end{equation}

where, $iqa\in\{SSIM,PSNR\}$.
To compute RA, we calculate the ratio of output images that correctly matched to their corresponding ground truth image, with respect to a biometric face matcher $\mathcal{B}$ and similarity threshold $\tau$ (typically set to 0.4\cite{ref66,ref18}),  to total number of output images. Assuming $(o_n,i_n), n\in\{1,2\}$ are matched, RA can be defined as
\begin{equation}
    RA=\frac{\sum\limits_{x\in \mathcal{X}}\mathbbm{1} \left( \mathcal{B}(o_1,i_1)>\tau \land  \mathcal{B}(o_2,i_2)>\tau\right)   }{ |\mathcal{X}|}
\end{equation}
where $|\mathcal{X}|$ is the number of morphs, $x=\mathcal{M}(i_1,i_2)$ and $o_1,o_2=\mathcal{DM}(x)$.


\subsection{Biometrically cross-weighted IQA}
To address the limitations of existing metrics (see Section \ref{sec:experiments}), we introduce a new metric: the biometrically cross-weighted IQA, defined as
\[
BW(iqa) = \mathbb{E}_{x \in \mathcal{X}} \max \left( 
\begin{aligned}
&\sum_{i\in\{1,2\}} \mathcal{B}(o_i, i_i) \cdot iqa(o_i, i_i), \\
&\sum_{\substack{i\in\{1,2\} \\ j=i\%2+1}} \mathcal{B}(o_i, i_{j}) \cdot iqa(o_i, i_{j}) 
\end{aligned}
\right)
\]  
where, $\%$ is the modulo operator and $iqa \in \{SSIM,PSNR\}$.  BW($iqa$) computes image quality between all possible  combinations of output and ground truth pairs and weighs them with the biometric match score. $max(\cdot)$ ensures that the correct pairing of output to ground truth is done during evaluation since the outputs are unordered during testing. 


\section{Datasets} 
\label{sec:datasets}
We train the methods under a common protocol, i.e., training is done with morphs created using synthetically generated faces and testing is done with morphs created using real faces. Note that this protocol is a more operationally viable scenario. 

\textbf{Train Dataset}: To train the models, we use the training face images from SMDD \cite{ref20} dataset.  The existing SMDD dataset is designed for Morph Attack Detection (MAD) and is not suitable for demorphing tasks. For example, the training set morphs in the SMDD dataset rely on only five
 images as the \textit{base image}, which can overfit the reconstruction of the second constituent image. Therefore, we generate morphs \textit{on-the-fly} for training.
During training, we randomly sample two face images and create a morph using the widely adopted \cite{ref88,sarkar2020vulnerabilityanalysisfacemorphing,9093905} OpenCV/dlib morphing algorithm \cite{ref97}, using Dlib's landmark detector implementation \cite{ref98}. We generate 15,000 train morphs and 15,000 test morphs using the training and test face images from SMDD dataset. All images (train and test) are processed using MTCNN \cite{ref71} to detect faces, after which the face regions are cropped. The images are then normalized and resized to a resolution of $256\times256$. Images where faces cannot be detected are discarded. Notably, no additional spatial transformations are applied, ensuring that the facial features (such as lips and nose) of both the morphs and the ground-truth constituent images remain aligned during training.

\textbf{Test Dataset}: We evaluate our proposed metric on three well-known morph datasets: AMSL \cite{ref64}, FRLL-Morphs \cite{ref65}, and MorDiff \cite{ref9}. The FRLL-Morphs dataset includes morphs generated using four different techniques: OpenCV \cite{ref97}, StyleGAN \cite{ref69}, WebMorph \cite{ref70}, and FaceMorph \cite{ref68}. In all three datasets, the source (i.e., non-morph) images come from the FRLL dataset, which includes 102 identities, each represented by two frontal images—one smiling and one neutral—resulting in a total of 204 bonafide images. The morph counts in each of the datasets are as follows: AMSL: 2,175 morphs; FaceMorpher: 1,222 morphs; StyleGAN: 1,222 morphs; OpenCV: 1,221 morphs; WebMorph: 1,221 morphs; MorDiff: 1,000 morphs. Note that the test morphs include those generated using  includes both conventional landmark-based techniques and more recently introduced generative methods.

\section{Experiments}
\label{sec:experiments}
\textbf{True Match Rate/RA is misleading: } Table \ref{tab:tmr} presents the True Match Rate @ 10\% False Match Rate across the three demorphing techniques evaluated on six datasets. While TMR is indicative of performance, a trivial solution, that simply replicates the morph image as its outputs, achieves perfect TMR, rendering the metric uninformative. By construction, a morph will have the highest similarity with the constituent images it is created from. Therefore, regurgitating the morph leads to the outputs matching with the constituent images with highest similarity, resulting in 100\% TMR. Restoration Accuracy (RA) is also limited by the same issue. A trivial solution leads to 100\% RA.  
 Moreover, TMR is computed in the score space (similarity score is calculated based on embeddings produced by a face matcher), which ignores the image quality in RGB space. 
 
\textbf{IQA (SSIM/PSNR):} Figure \ref{fig:ssim-psnr} illustrates the failure cases of SSIM and PSNR when used for facial demorphing. In the most severe case (Figure \ref{fig:ssim-psnr}, second row), the SSIM and PSNR values of image $\mathcal{I}_1$  with $\mathcal{I}_2$ indicate  that $\mathcal{I}_1$ is structurally closer to $\mathcal{I}_2$ - a face image belonging to different identity - compared to a noisy version of the same face image. This indicates an important issue with metrics that operate in RGB pixel space; they ignore the identity information, making them unsuitable for demorphing tasks. Indeed, in Table \ref{tab:results}, \cite{ref51} achieves superior results in terms of PSNR and SSIM despite the output images looking significantly distorted compared to \cite{ref18} and \cite{ref66}. Our proposed metric balances the structural similarity in RGB pixel space with the biometric similarity computed in feature space to produce score metrics consistent with the visual inspection of the outputs.
\section{Results}
\label{sec:results}
We benchmark three existing demorphing methods under a unified protocol: Facial Demorphing \cite{ref51}, SDeMorph \cite{ref18}, and Identity Preserving Decomposition (IPD) \cite{ref66}. These methods are evaluated using both established metrics and the proposed metric. To compute similarity scores, we employ two widely used face matchers, AdaFace \cite{ref22} and ArcFace \cite{ref77}. Table \ref{tab:tmr} presents the TMR @ 10\% FMR across six datasets, averaged over two subjects. While the results provide insights into performance—e.g., IPD achieves the highest TMR, aligning with visual inspection—the metric is rendered uninformative when compared to a trivial solution that achieves perfect TMR. We also compute SSIM/PSNR and Restoration Accuracy and present the results in Table \ref{tab:results}. \cite{ref51} achieves the best performance in terms of Restoration Accuracy and PSNR/SSIM despite having more visible high frequency artifacts (see Figure \ref{fig:enter-label}) indicating the vulnerability when used for demorphing tasks. Moreover, RA suffers the same issue as TMR since the trivial solution leads to perfect results. Finally, we present $BW(PSNR)$ and $BW(SSIM)$ in Table \ref{tab:results}. SDeMorph achieves $BW(SSIM)$ of 0.14 and $BW(PSNR)$ of 3.62, averaged across the two face matchers. These scores for IPD and Facial Demorphing are 0.23/6.0 and 0.17/3.84, respectively. IPD\cite{ref66} performs the best among the three methods tested in terms of the proposed metric. This is also consistent with the visual results produced by the method indicating the relevance of the metric for demorphing tasks.  

\section{Conclusion}
We propose a new evaluation metric for reference-free face demorphing methods. Existing metrics either focus solely on biometric features and ignore image quality (TMR and RA) or completely ignore the biometric aspect (SSIM/PSNR). Moreover, by construction, the morph will have highest similarity with the constituent images used to create it. This results in misleading TMR and RA scores. On the other hand, Image Quality Assessment (IQA) metrics like SSIM/PSNR operate in RGB pixel space and do not capture the biometric information completely. Our proposed metric weights and combines IQA as well as biometric similarity and produces consistent scores across the experiments conducted. 

\label{sec:conclusion}
{\small
\bibliographystyle{plainurl}
\bibliography{egbib}
}

\end{document}